\documentclass[10pt,twocolumn,letterpaper]{article}

\usepackage{cvpr}              

\usepackage{graphicx}
\usepackage{amsmath}
\usepackage{amssymb}
\usepackage{booktabs}
\usepackage{multirow}
\usepackage{array}

\usepackage[pagebackref,breaklinks,colorlinks]{hyperref}

\usepackage[capitalize]{cleveref}
\crefname{section}{Sec.}{Secs.}
\Crefname{section}{Section}{Sections}
\Crefname{table}{Table}{Tables}
\crefname{table}{Tab.}{Tabs.}

\def\cvprPaperID{*****} 
\def\confName{CVPR}
\def\confYear{2022}

\begin{document}

\title{Explainability of Deep Neural Networks for Brain Tumor Detection}

\author{Sunyoung Park\thanks{Corresponding author: tjsdud9151@g.skku.edu}\\
Department of Applied Artificial Intelligence\\
Sungkyunkwan University\\
Seoul 03063, Korea\\
{\tt\small tjsdud9151@g.skku.edu}
\and
Jihye Kim\\
Department of Applied Artificial Intelligence\\
Sungkyunkwan University\\
Seoul 03063, Korea\\
{\tt\small kcom0712@gmail.com}
}
\maketitle

\begin{abstract}
   Medical image classification is crucial for supporting healthcare professionals in decision-making and training. While Convolutional Neural Networks (CNNs) have traditionally dominated this field, Transformer-based models are gaining attention. In this study, we apply explainable AI (XAI) techniques to assess the performance of various models on real-world medical data and identify areas for improvement. We compare CNN models such as VGG-16, ResNet-50, and EfficientNetV2L with a Transformer model: ViT-Base-16. Our results show that data augmentation has little impact, but hyperparameter tuning and advanced modeling improve performance. CNNs, particularly VGG-16 and ResNet-50, outperform ViT-Base-16 and EfficientNetV2L, likely due to underfitting from limited data. XAI methods like LIME and SHAP further reveal that better-performing models visualize tumors more effectively. These findings suggest that CNNs with shallower architectures are more effective for small datasets and can support medical decision-making.
\end{abstract}

\section{Introduction}
Brain tumors, originating within the cranium, manifest as malignant or benign neoplasms, inflicting detrimental effects on neural functionalities and precipitating a plethora of symptoms. They are underscored as one of the pivotal causatives of adult mortality. However, early detection paired with proactive therapeutic interventions amplifies the potential to substantially augment survival rates. The diagnostic and therapeutic stratagems pivoting around brain tumors are integrally reliant on medical imaging modalities, predominantly Magnetic Resonance Imaging (MRI). The conventional methodologies, such as iterative surgical evaluations, are not feasible for assessing therapeutic responses due to the inherent complexities in visually delineating the tumor from the surrounding brain parenchyma. Thus, MRI emerges as an indispensable tool, facilitating a visual encapsulation of the cerebral architecture, tumor localization and dimension, as well as histological characteristics\cite{tu2021gold}. In alignment with the advancements in artificial intelligence, deep learning has burgeoned as a profoundly promising instrument in the realm of brain tumor detection. Nevertheless, the deep learning techniques employed in the majority of precedent studies possess an intrinsic limitation — they proffer effective decision-support mechanisms but are simultaneously enshrouded in algorithmic complexities that obfuscate the interpretability of the derived outcomes\cite{zhu2021interpreting}. Particularly in the medical domain, where a paramount emphasis is placed on reliability and accuracy, reliance solely on result values engenders substantial risks. Consequently, there emerges a pronounced necessity for technologies that foster accurate estimations and interpretations, tailoring to the nuanced requirements intrinsic to the healthcare sector. In response to this need, there has been a growing interest in XAI research to detect brain tumors using artificial intelligence models and to identify the reasons for the models' decisions. In this study, we compare the performance of brain tumor detection models using CNN and Transformer-based pre-trained models that have reported good performance in image classification, and introduce image augmentation and parameter tuning methods for further improvement. In addition, by explaining the results using three different methodologies, LIME, SHAP, and Grad-CAM, it will be possible to effectively deal with the limitations of current diagnostics and provide interpretable information about the results. 

\section{Related work}
Medical image classification is a sub-subject of image classification\cite{yadav2019deep}. This section synthesizes efforts in image classification and explainability in the medical domain and introduces related techniques.
\subsection{Deep learning and XAI in healthcare}
Yadav \& Jadhav \cite{yadav2019deep} use VGG16 and InceptionV3 for pneumonia classification, performing transfer learning and capsule network training, and checking their effectiveness against data augmentation. Yan et al.\cite{yan2021neural} use VGG-16 and ResNet to predict glioblastoma progression phenotypes using MRI data. Then, using CAM, they visualize the neural network to identify localized and diffuse patterns.Kim and Ye\cite{kim2020understanding} analyzed fMRI data using Graph Isomorphism among graph neural networks, and visualized important regions of the brain using CNN-based saliency map for GNNs.
Kubach et al. \cite{kubach2020same} used CNNs to distinguish between focal cortical dysplasia type IIb and cortical tuber of tuberous sclerosis complex, and Guided Grad-CAM to visualize morphological features that distinguish the two entities. These patterns were then combined into a classification score, validated on subjects consisting of experts and non-experts, and claimed that non-experts can make decisions at an expert level.
Araújo et al.\cite{araujo2020dr} proposed a novel deep learning approach for diabetic retinopathy (DR) grading, DR|GRADUATE, which used multiple instance learning to impose explainability, which had the advantage of generating both an estimate of the uncertainty associated with the prediction and an explanatory map highlighting the most relevant regions for classification, without degrading DR grading performance.
Hossain et al. \cite{hossain2023vision} compare the performance of CNN-based models (VGG-16, InceptionV3, ResNet50, Xception, VGG-19, InceptionResNetV2) for Brain Tumor Detection and propose an ensemble of them, IVX16. They also visualize them using LIME.
Zhu \& Ogino\cite{zhu2019guideline} proposed a Guideline-based Additive eXplanation (GAX) framework for diagnosing pulmonary nodules. They compared their method with LIME and SHAP, and claimed that their results were similar to LIME and SHAP, but they increased the anatomical meaning. In these various fields, AI is used to perform disease classification, and XAI is added to provide explainability to improve the understanding of medical staff and general users. We refer to these studies to develop a disease prediction model based on deep learning and apply LIME for explainability.\\
By applying AI to the medical field, these studies can help medical staff make decisions and automate diagnosis. They also provide clues to solve the black box problem of AI by showing the evidence to medical staff. Nevertheless, most of the studies\cite{yadav2019deep, kim2020understanding, yan2021neural, yadav2019deep} focus on developing AI models or improving their performance through hyperparameters and image preprocessing, so they only partially represent the decision-making process of the model rather than visualizing it using multiple explanatory AI techniques. In addition, Zhu \& Ogino\cite{zhu2019guideline} compare various visualization methods and present a framework, but training and experimentation on real data are not performed, leaving some gaps between modeling and explainable techniques. Therefore, in this study, rather than proposing a model that performs better than previous studies, we perform transfer learning on multiple models to achieve robust performance without a separate modeling process and compare their performance. We also provide experimental results on image preprocessing and parameter tuning to improve them, and use three representative explainable AI techniques, LIME, SHAP, and Grad-CAM, to visualize them and provide explanatory rationales for the models.

\subsection{Classification Models}
\subsubsection{VGG-16 architecture}
\begin{figure}[ht]
  \centering
  \includegraphics[width=0.45\textwidth]{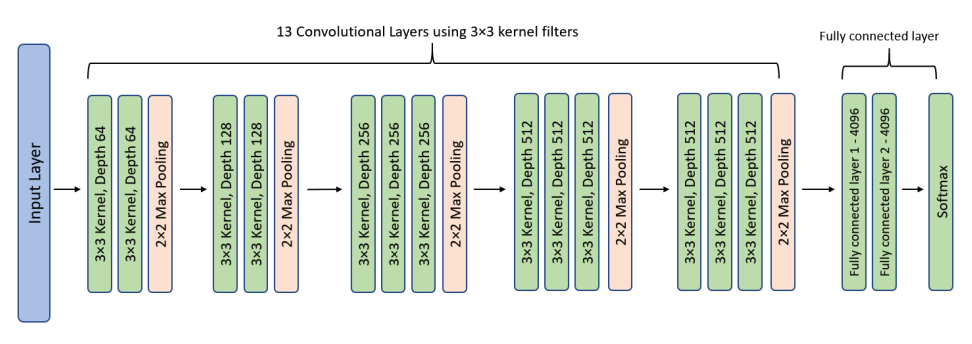}
  \caption{Block diagram of VGG-16 network \cite{tammina2019transfer}}
  \label{fig:vgg}
\end{figure}

VGG-16 is a model that performed well in the ImageNet Challenge 2014, the VGG-16 network consists of 16 convolutional layers and has a small receptive field of 3 × 3 \cite{simonyan2014very}. The max pooling layer of this model has a size of 2 × 2, there are 5 layers in total, after the last max pooling layer there are three fully connected layers. After that there are three fully connected layers, and a softmax classifier is used as the final layer. ReLu activation is applied to all hidden layers. The schematic of the VGG-16 architecture can be seen in Figure \ref{fig:vgg}.

\subsubsection{ResNet-50 architecture}
\begin{figure}[ht]
  \centering
  \includegraphics[width=0.45\textwidth]{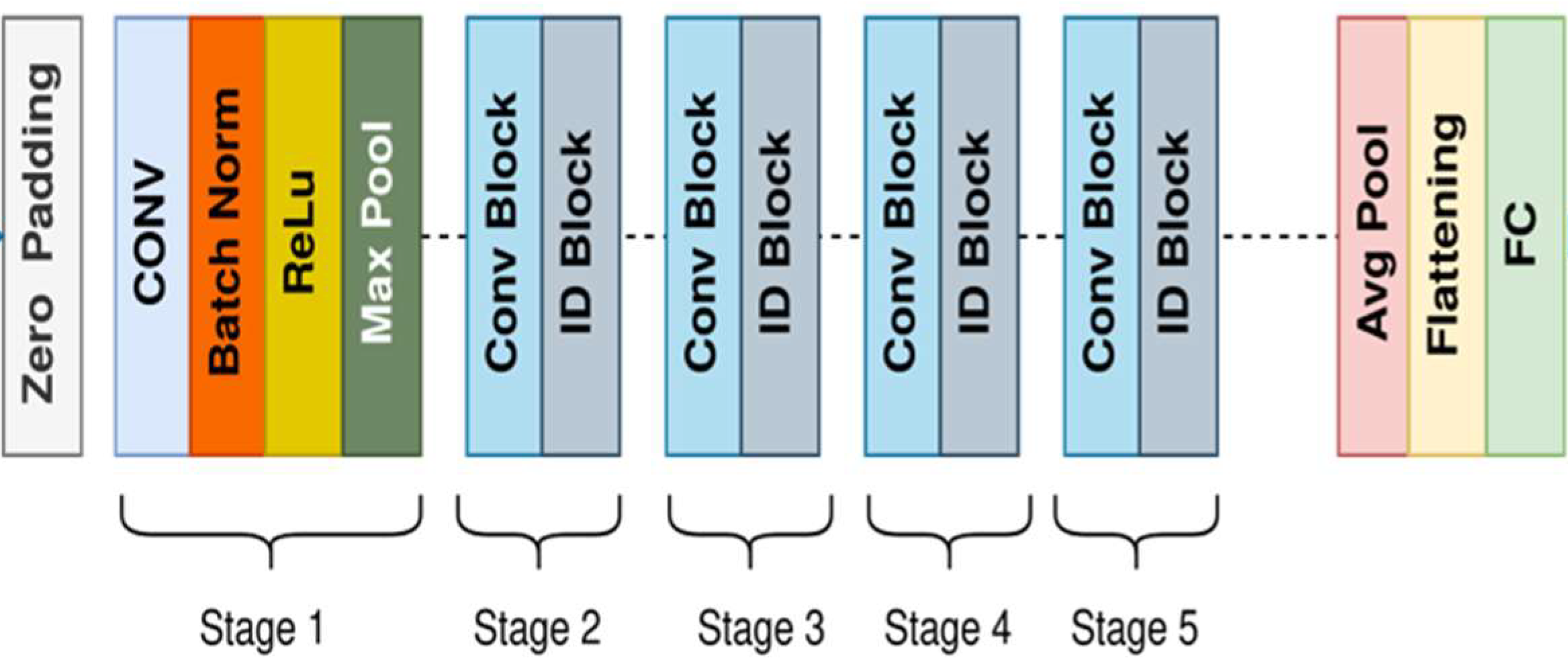}
  \caption{ResNet-50 architecture\cite{shatnawi2023deep}}
  \label{fig:resnet_50}
\end{figure}

ResNet50 (Figure \ref{fig:resnet_50}) is a shorthand for residual networks with 50 layers, where the model predicts the delta needed to progress from one layer to the next \cite{he2016deep}. ResNet addresses the vanishing gradient problem by introducing an additional shortcut path for the gradient to flow through \cite{theckedath2020detecting}. The use of identity mapping in ResNet enables the model to skip a CNN weight layer if the current layer is deemed unnecessary, thereby helping to mitigate overfitting issues in the training set. It's worth noting that ResNet50 specifically consists of 50 layers \cite{theckedath2020detecting}.

\subsubsection{EfficientNet architecture}
\begin{figure}[ht]
  \centering
  \includegraphics[width=0.45\textwidth]{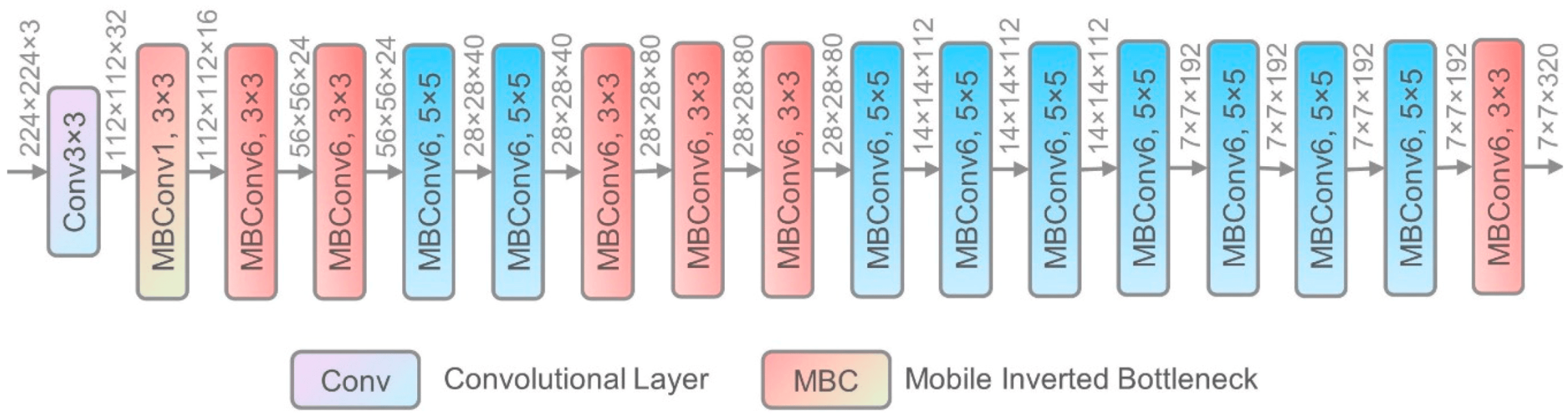}
  \caption{EfficientNet architecture\cite{xu2021forest}}
  \label{fig:eff}
\end{figure}

EfficientNet (Figure \ref{fig:eff}) is a new efficient network proposed by Google \cite{xu2021forest}. It applied a novel model scaling strategy, namely compound scaling method, to balance network depth, network width, and image resolution for better accuracy at a fixed resource budget \cite{xu2021forest}. The last layer of the EfficientNet-V2L is replaced and 2-dimensional global average pooling is applied\cite{joshi2022forged}. Followed by it is one dense layer with 1024 hidden neurons and ReLU activation function. The final output layer is applied with 2 hidden neurons and softmax activation function. The loss function used is binary cross-entropy and adam loss optimizer for loss optimization.

\subsubsection{ViT architecture}
\begin{figure}[ht]
  \centering
  \includegraphics[width=0.45\textwidth]{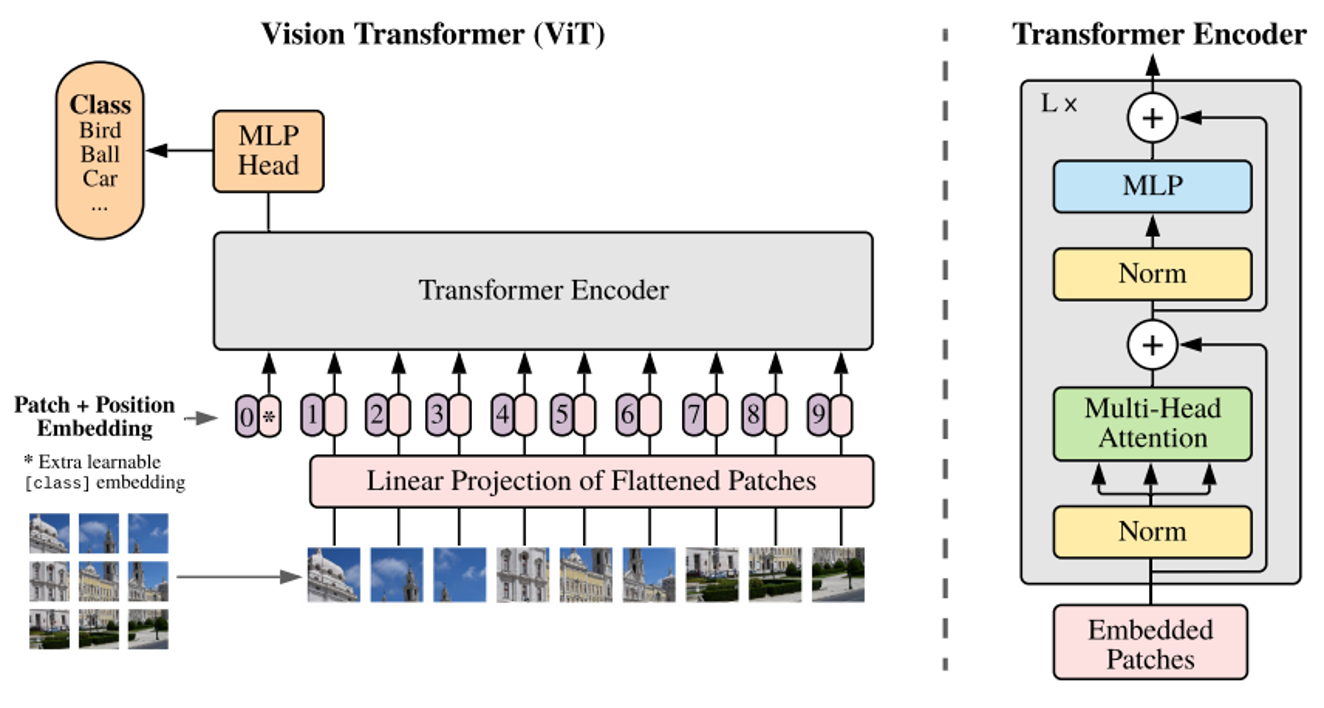}
  \caption{ViT architecture\cite{dosovitskiy2020image}}
  \label{fig:vit}
\end{figure}

ViT is a deep learning architecture for computer vision tasks that applies the Transformer structure to image processing\cite{dosovitskiy2020image}. ViT segments an image into a series of patches, which are then converted into one-dimensional vectors that are used as input to the Transformer. These transformed patches are then fed into the Transformer's encoder, which is used to recognize the image by modeling the interrelationships of the input patches. ViT's encoder has almost the same structure as the original Transformer, with the main difference being that the input is a one-dimensional vector, and ViT uses patch embedding to convert the input patches into vectors. ViT also uses positional embedding to add the positional information of the patches to the input of the Transformer. A schematic of the ViT architecture can be seen in Figure \ref{fig:vit}.

\subsection{Explainable AI Models}
\subsubsection{LIME}
\begin{figure}[ht]
  \centering
  \includegraphics[width=0.35\textwidth]{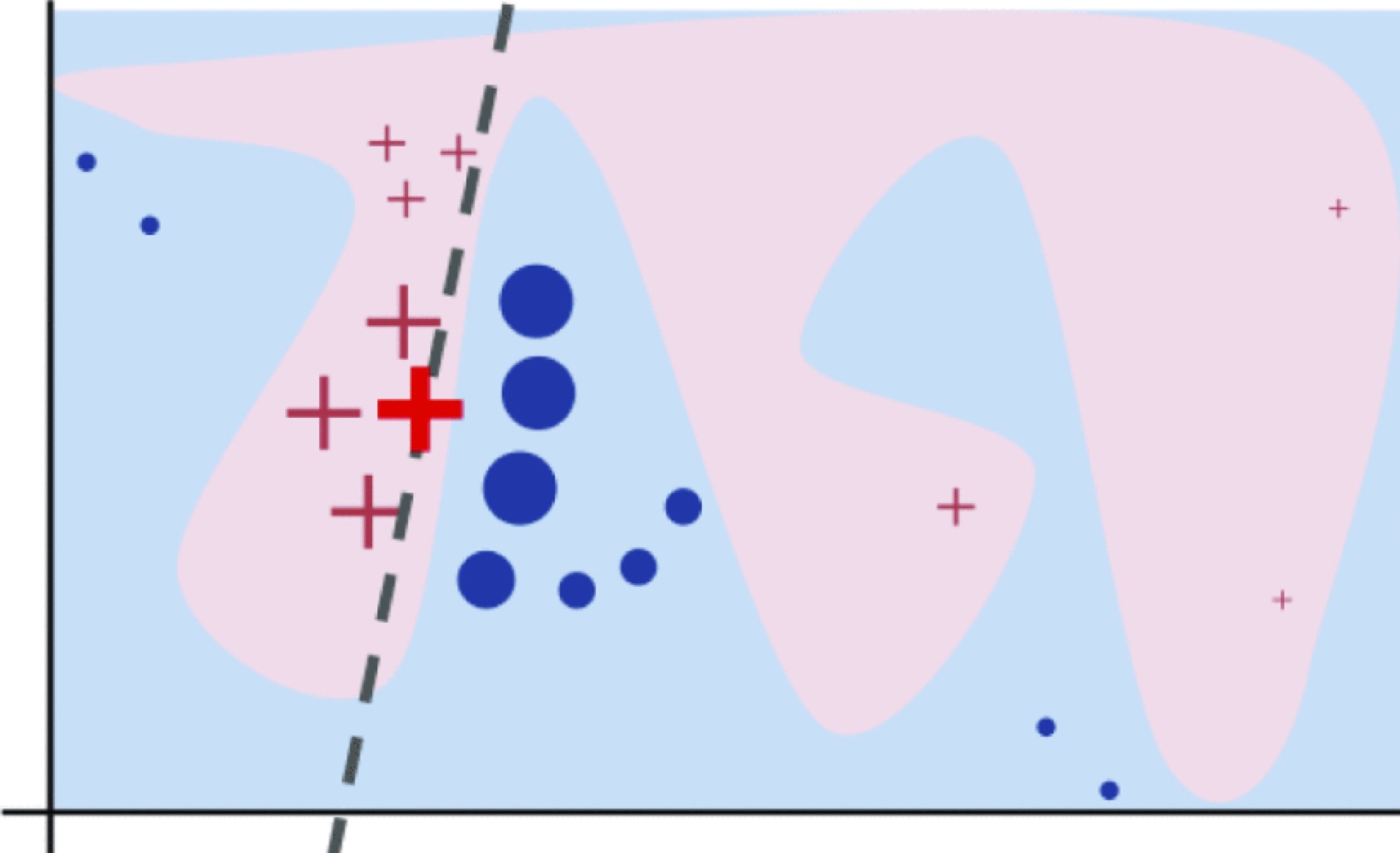}
  \caption{Explaining the predictions using LIME\cite{ribeiro2016should}}
  \label{fig:LIME}
\end{figure}

LIME is an XAI method that can explain the predictions of any classifier or regressor reliably by approximating it locally with an interpretable model\cite{ribeiro2016should}. This is a model-agnostic method because it can describe any model without having to investigate the model. For classification, LIME generates a dataset by perturbing the original test instances, and for each perturbed sample, it calculates the probability that the test instance belongs to a particular class\cite{panati2022feature}. Figure \ref{fig:LIME} shows how Lime fits a local linear model for the selected forecast.

\subsubsection{SHAP}
SHAP \cite{lundberg2017unified} method, which is a unified approach that aims to explain the model output using shapely values - a concept borrowed from game theory where it is used to calculate the relative contributions of different players in a coalition. In the context of XAI, they are used for estimating the contribution of a specific input or neuron to a model’s decision\cite{fidel2020explainability}.

\subsubsection{Grad-CAM}
\begin{figure}[ht]
  \centering
  \includegraphics[width=0.45\textwidth]{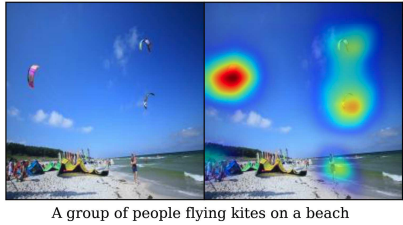}
  \caption{Explaining the predictions using LIME\cite{selvaraju2017grad}}
  \label{fig:Grad-CAM}
\end{figure}

Grad-CAM\cite{selvaraju2017grad} is a technique for making CNN-based models more transparent by creating a 'visual' explanation of their decisions, using the gradient of the target concept to flow into the final convolution layer to create a localization map that highlights areas that are important when predicting the concept. An example visualization using Grad-CAM is shown in Figure \ref{fig:Grad-CAM}.

\section{Method}
\subsection{Data}
\begin{figure}[ht]
  \centering
  \includegraphics[width=0.35\textwidth]{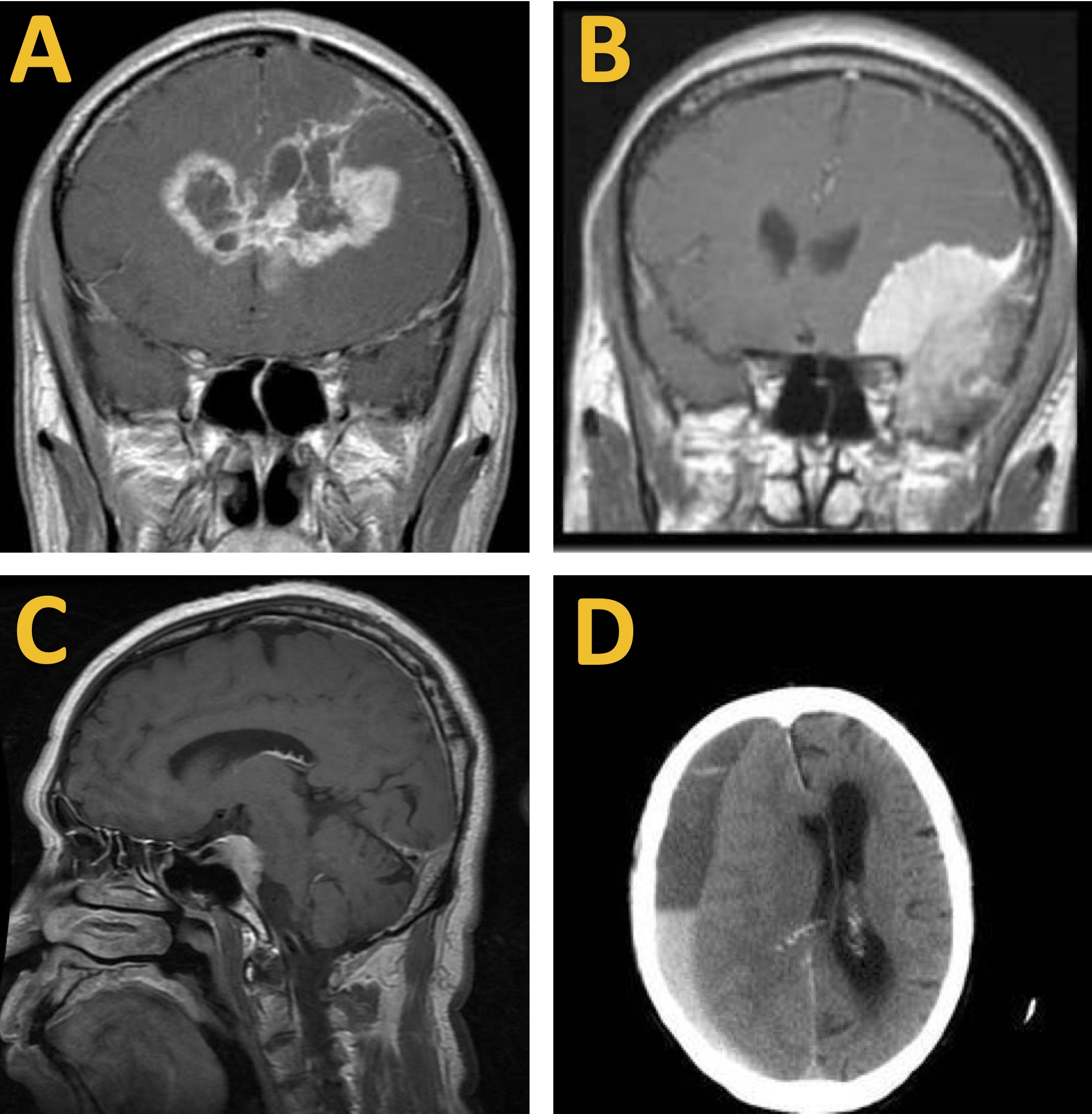}
  \captionsetup{justification=centering}
  \caption{Example of Dataset\\(A: Glioma, B: Meningioma, C: Pituitary, D: No Tumor)}
  \label{fig:data}
\end{figure}

In this paper, we train our model using Brain Tumor Classification\cite{grover2023braintumor}, an open dataset on Kaggle. The code used in this study can be found on GitHub \url{https://github.com/sunyoung98/Brain_Tumor_Detection_XAI}. This dataset aims to classify four types of brain tumors, which include Glioma, Meningioma, Pituitary and No tumor. The dataset contains a total of 3,264 images and consists of Glioma: 926, Meningioma: 837, Pituitary: 901, and No tumor: 500. In the training phase, we equalize the amount of images used to train the model for each class or tumor type. Out of all the available images, in each class, 80\% is used for training, 10\% for validation, and 10\% is reserved for testing the model. The datasets were provided in .jpg format. An example of a simple dataset is shown below (Figure \ref{fig:data}).\\

The image preprocessing would follow the criteria specified in the pretraining model, and the criteria for both models were the same: each image is resized to 256 x 256, with BILINEAR used as the interpolation method. Next, they are centered cropped to a size of 224 x 224 and scaled to a value between [0.0, 1.0]. Next, it is normalized with mean=[0.485, 0.456, 0.406], std=[0.229, 0.224, 0.225]. 

\subsection{Transfer Learning Experiments Design}
We compare the performance of the four models considered in this study after transfer learning. We let them learn 100 epochs, and the best model is determined based on the validation loss. VGG-16, ResNet-50, and EfficientNetV2L are pre-trained models in Keras, while ViT/B-16 is pre-trained in Keras\_vit. All models are pre-trained on ImageNet, and after flattening the output of the model, we add a dense layer with an activation function of softmax. The optimizer for all models is SGD, the learning rate is 0.001, and the loss is calculated as categorical cross entropy.

\subsection{Data Augmentation Design}
In this study, we compare two data augmentation algorithms. We followed the augmentation method of Yadav \& Jadhav \cite{yadav2019deep}, a related prior work. As shown in Table \ref{tab:augmentation}, Aug1 refers to simple geometric transformations of the image. For example, it randomly flips horizontally and vertically, rotates randomly within 0.05, and horizontal shears within 0.05 times the width of the image, while Aug2 is a more complex transformation than Aug1. In addition to all of the transformations in Aug1, it also performs some horizontal and physical translation.\\

\begin{table}[ht]
\centering
\caption{Augmentation Models}
\begin{tabular}{p{2cm} p{5.5cm}}
\toprule
\multicolumn{1}{c}{\textbf{Augmentation}} & \multicolumn{1}{c}{\textbf{Parameters}} \\
\midrule
\multicolumn{1}{c}{\textbf{Aug1}} & Rotation range = 0.05, shear range = 0.05, zoom range = 0.05, horizontal flip = True, vertical flip = True \\
\multicolumn{1}{c}{\textbf{Aug2}} & Rotation range = 3, width shift range = 0.05, height shift range = 0.05, shear range = 0.05, zoom range = 0.05, fill mode = 'constant', cval = 0, horizontal flip = True, vertical flip = True \\
\bottomrule
\end{tabular}
\label{tab:augmentation}
\end{table}

\subsection{Parameter Tuning Experiments Design}
Hyper parameter tuning experiments with the effects on learning rate, optimizer, and epochs. Hyper parameter tuning is partially experimented on two models that reported good performance during the transfer learning experiments. Learning rate is experimented with three settings: 1) 0.001, 2) 0.005, and 3) 0.01. optimizer is experimented with two settings: 1) SGD and 2) Adam. The effect of epochs is experimented with three settings: 1) 10 epochs, 2) 20 epochs, and 3) 50 epochs. The overall experimental conditions were as follows (table \ref{tab:parameter}).\\
\begin{table}[ht]
\centering
\caption{Parameter Tuning Experiments Design}
\label{tab:parameter}
\begin{tabular}{p{3cm} p{4cm}}
\toprule
\multicolumn{1}{c}{\textbf{Parameter}}       & \multicolumn{1}{c}{\textbf{Condition}} \\ 
\midrule
\multicolumn{1}{c}{Learning Rate}   & 1) 0.001, 2) 0.005, 3) 0.01              \\ 
\multicolumn{1}{c}{Optimizer}       & 1) SGD, 2) Adam                          \\ 
\multicolumn{1}{c}{Training Epochs} & 1) 10, 2) 20, 3) 50 epochs               \\ 
\bottomrule
\end{tabular}
\end{table}

\subsection{Additional Modeling Experiments Design}
To conduct our experiments, we set up the model in Section 3.1 with the classification layer added and the previous layers frozen. This is a common practice in transfer learning models, where only one layer of the classification layer is trained with weights, which has the effect of controlling the number of training parameters for the entire model while allowing for updates to specific layers. However, this may not be enough to perform classification, since the first layer is the only one that can be trained. Therefore, we leave the previous layers frozen and allow training to occur only for the newly added classification layer.\\
We experimented with a total of 8 cases (Table \ref{tab:model_design}) after the output layers of VGG-16 and ResNet-50, we added dense layers in the following order: 256, 128, 64, 32, and then 30\% dropout. All training parameters were used as described in Section 3.1.\\

\begin{table}[ht]
\centering
\caption{Model Design}
\label{tab:model_design}
\begin{tabular}{>{\centering\arraybackslash}p{2cm} p{5.5cm}}
\toprule
\multicolumn{1}{c}{\textbf{Case}} & \multicolumn{1}{c}{\textbf{Model Design}} \\ 
\midrule
\textbf{Version 1} & 1-layer: Dense(256) \\ 
\textbf{Version 2} & Version 1 + Dropout = 0.3 \\ 
\textbf{Version 3} & 2-layer: Dense(256), Dense(128) \\ 
\textbf{Version 4} & Version 3 + Dropout = 0.3 \\ 
\textbf{Version 5} & 3-layer: Dense(256), Dense(128), Dense(64) \\ 
\textbf{Version 6} & Version 5 + Dropout = 0.3 \\ 
\textbf{Version 7} & 4-layer: Dense(256), Dense(128), Dense(64), Dense(32) \\ 
\textbf{Version 8} & Version 7 + Dropout = 0.3 \\ 
\bottomrule
\end{tabular}
\end{table}

\section{Experiment Results of Classification}
\subsection{Transfer Learning Classification}
Before we look at the Transfer Learning results for the four models, we additionally check the results trained on an untrained model (Table \ref{tab:performance}. This is a reflection of the professor's feedback during the presentation. The pre-trained model had the same hyperparameter settings as Transfer Learning, with the epoch adjusted to 20. The reason why the epoch was adjusted to 20 is that it took much longer to learn because of the large number of parameters that needed to be learned, and especially in the case of efficientNet, the network size is large, and all the parameters and settings required for the computation process must be uploaded to the memory, so there was an out of memory problem, so it was not possible to even proceed with the learning with the resources we had, and it took about 2 hours per epoch in a progressive environment. In particular, when running experiments in a local environment, the model was not trained properly (accuracy/loss barely changed in the same code). Because of this very high computing power requirement without transfer learning, we only trained 20 epochs. The experimental results show very poor performance for VGG-16, comparable results for ResNet, and better performance for ViT than when using a pre-trained model. 
\begin{table}[ht]
\centering
\caption{Performance of Not-pretrained Models}
\label{tab:performance}
\begin{tabular}{l c c c}
\toprule
          & \textbf{ResNet} & \textbf{VGG-16} & \textbf{ViT}   \\ 
\midrule
\textbf{Precision} & 0.847  & 0.345 & 0.912 \\ 
\textbf{Recall}    & 0.803  & 0.372 & 0.906 \\ 
\textbf{F1-score}  & 0.808  & 0.284 & 0.907 \\ 
\textbf{Accuracy}  & 0.819  & 0.390 & 0.908 \\ 
\bottomrule
\end{tabular}
\end{table}

\begin{figure}[ht]
  \centering
  \includegraphics[width=0.45\textwidth]{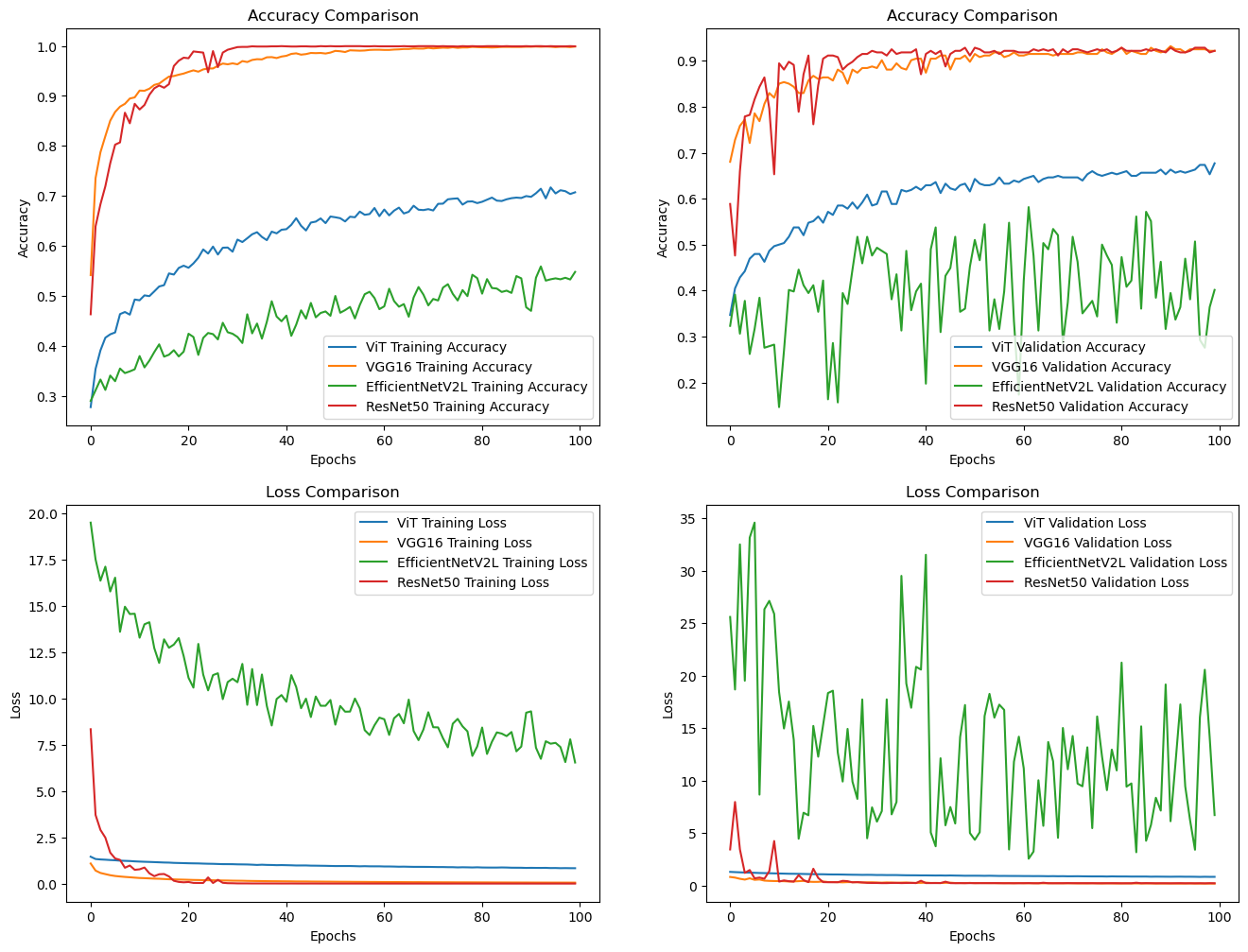}
  \captionsetup{justification=centering}
  \caption{Performance of Transfer Learning Models\\ (red: ResNet, orange: VGG, green: EfficientNet, blue: ViT)}
  \label{fig:pre-trained}
\end{figure}

The transfer learning results for the four models are shown below. The loss and accuracy of training and validation of all models can be seen in Figure \ref{fig:pre-trained} and Table \ref{tab:pre-trained}. VGG-16 and ResNet-50 report similarly high performance, with VGG-16 learning more reliably. ViT/B-16 and EfficientNet are next, with similar performance on our test set.

\begin{table}[ht]
\centering
\caption{Performance of Transfer Learning Models}
\label{tab:pre-trained}
\begin{tabular}{l c c c c}
\toprule
          & \textbf{ResNet} & \textbf{VGG}   & \textbf{EfficientNet} & \textbf{ViT}   \\ 
\midrule
\textbf{Precision} & 0.924  & \textbf{0.932} & 0.670        & 0.665 \\ 
\textbf{Recall}    & 0.920  & \textbf{0.927} & 0.597        & 0.649 \\ 
\textbf{F1-score}  & 0.921  & \textbf{0.927} & 0.579        & 0.647 \\ 
\textbf{Accuracy}  & 0.920  & \textbf{0.927} & 0.587        & 0.654 \\ 
\bottomrule
\end{tabular}
\end{table}

\subsection{The Effect of Data Augmentation}
The Table \ref{tab:Augmentation_Result} summarizes the results of our experiments using three different augmentations. We can see that augmentations have little effect regardless of the model. In particular, the more complex the transformation, the worse the performance, so it may be better to experiment further to find the right augmentation settings, or to use the original data without augmentation. This suggests that more complex transformations introduce noise that interferes with the learning process.

\begin{table}[ht]
\centering
\caption{The Comparison of Data Augmentation Experiments}
\label{tab:Augmentation_Result}
\begin{tabular}{c c c c c c}
\toprule
                      & \textbf{Model}     & \textbf{Precision} & \textbf{Recall} & \textbf{F1} & \textbf{Acc} \\ 
\midrule
\multirow{4}{*}{\textbf{Aug1}} & ResNet    & \textbf{0.915}     & \textbf{0.913}  & \textbf{0.914}    & \textbf{0.914}    \\ 
                      & VGG       & 0.907     & 0.903  & 0.905    & 0.905    \\ 
                      & Efficient & 0.610     & 0.513  & 0.528    & 0.514    \\ 
                      & ViT       & 0.651     & 0.631  & 0.628    & 0.636    \\ 
\midrule
\multirow{4}{*}{\textbf{Aug2}} & ResNet    & 0.862     & 0.852  & 0.855    & 0.853    \\ 
                      & VGG       & \textbf{0.919}     & \textbf{0.917}  & \textbf{0.917}    & \textbf{0.917}    \\ 
                      & Efficient & 0.582     & 0.493  & 0.474    & 0.505    \\ 
                      & ViT       & 0.645     & 0.625  & 0.623    & 0.630    \\ 
\bottomrule
\end{tabular}
\end{table}

\subsection{The Effect of Hyper Parameter Tuning}
The full results of the hyperparameter tuning can be found in the Table \ref{tab:Hyper_Parameter}. In the main text, we only show the performance for the optimal parameter combinations. We further experimented with ResNet-50 and VGG-16, which had high performance, and the optimal parameter combinations are shown in bold.\\
For ResNet, the model trained with Learning rate: 0.005, Optimizer: SGD, 20 Epochs performs the best. For training epochs, the validation loss may be lower when training more epochs, but it may be due to overfitting on the validation set and not showing robust performance on the test set. For VGG-16, it performs best with Learning rate: 0.01, Optimizer: SGD, 20 Epochs, and the lack of performance improvement when increasing the number of training epochs may be due to the same reason as ResNet.
\begin{table}[ht]
\centering
\caption{Effects of Hyper Parameter Tuning (Accuracy)}
\label{tab:Hyper_Parameter}
\begin{tabular}{l l c c c}
\toprule
\multicolumn{5}{c}{\textbf{ResNet-50}}  \\ 
\midrule
\textbf{LR} & \textbf{Optim} & \textbf{10 Epoch} & \textbf{20 Epoch} & \textbf{50 Epoch} \\ 
\midrule
\multirow{2}{*}{0.001} & SGD  & 0.88 & 0.81 & 0.92 \\ 
                       & Adam & 0.92 & 0.88 & 0.91 \\ 
\midrule
\multirow{2}{*}{0.005} & SGD  & 0.85 & \textbf{0.93} & 0.91 \\ 
                       & Adam & 0.83 & 0.85 & 0.90 \\ 
\midrule
\multirow{2}{*}{0.01}  & SGD  & 0.81 & 0.89 & 0.90 \\ 
                       & Adam & 0.88 & 0.90 & 0.91 \\ 
\midrule
\multicolumn{5}{c}{\textbf{VGG-16}} \\ 
\midrule
\textbf{LR} & \textbf{Optim} & \textbf{10 Epoch} & \textbf{20 Epoch} & \textbf{50 Epoch} \\ 
\midrule
\multirow{2}{*}{0.001} & SGD  & 0.88 & 0.90 & 0.91 \\ 
                       & Adam & 0.94 & 0.94 & 0.94 \\ 
\midrule
\multirow{2}{*}{0.005} & SGD  & 0.87 & 0.93 & 0.94 \\ 
                       & Adam & 0.94 & 0.92 & 0.92 \\ 
\midrule
\multirow{2}{*}{0.01}  & SGD  & 0.90 & \textbf{0.96} & 0.94 \\ 
                       & Adam & 0.91 & 0.92 & 0.93 \\ 
\bottomrule
\end{tabular}
\vspace{5pt}
\caption*{LR refers to learning rate and Optim refers to optimizer.}
\end{table}

\subsection{Effects of Adding Layers}
You can see all the results of the experiments to see the effect of additional layers in Table \ref{tab:Adding_Layers}. The following parameters are used: Learning rate: 0.01, Optimizer: SGD, Epochs: 50. The best performance is reported by Version, which adds one layer of Dense Layer and 30\% dropout. The reason for the higher performance compared to adding only the classification layer is that by adding the Desne Layer, it helps to learn the features of the input data, and by adding the dropout, it prevents overfitting and increases the generalization performance of the model. Although most of the models report better performance than the original model, for Vgg-16, Version 5 reports lower performance than the original. Therefore, depending on the characteristics of the dataset and the problem to be solved, it is recommended to select a model design and conduct further experiments to determine the optimal model.\\

\begin{table}[ht]
\centering
\caption{Effects of Adding Layers}
\label{tab:Adding_Layers}
\begin{tabular}{>{\centering\arraybackslash}p{1.5cm} c c c c}
\toprule
\multirow{2}{*}{} & \multicolumn{4}{c}{\textbf{ResNet-50}} \\ 
\cmidrule{2-5}
                  & \textbf{Precision} & \textbf{Recall} & \textbf{F1-score} & \textbf{Acc} \\ 
\midrule
\textbf{Original} & 0.924  & 0.920  & 0.921  & 0.920  \\ 
\textbf{Version 1}    & 0.942  & 0.939  & 0.939  & 0.939  \\ 
\textbf{Version 2}    & \textbf{0.952}  & \textbf{0.951}  & \textbf{0.951}  & \textbf{0.951}  \\ 
\textbf{Version 3}    & 0.937  & 0.936  & 0.936  & 0.936  \\ 
\textbf{Version 4}    & 0.928  & 0.927  & 0.927  & 0.927  \\ 
\textbf{Version 5}    & 0.945  & 0.945  & 0.945  & 0.945  \\ 
\textbf{Version 6}    & 0.928  & 0.926  & 0.926  & 0.927  \\ 
\textbf{Version 7}    & 0.944  & 0.942  & 0.942  & 0.942  \\ 
\textbf{Version 8}    & 0.940  & 0.938  & 0.939  & 0.939  \\ 
\midrule
\multirow{2}{*}{} & \multicolumn{4}{c}{\textbf{VGG-16}} \\ 
\cmidrule{2-5}
                  & \textbf{Precision} & \textbf{Recall} & \textbf{F1-score} & \textbf{Acc} \\ 
\midrule
\textbf{Original} & 0.932  & 0.927  & 0.927  & 0.927  \\ 
\textbf{Version 1}    & 0.944  & 0.941  & 0.942  & 0.942  \\ 
\textbf{Version 2}    & \textbf{0.947}  & \textbf{0.945}  & \textbf{0.945}  & \textbf{0.945}  \\ 
\textbf{Version 3}    & 0.934  & 0.927  & 0.927  & 0.927  \\ 
\textbf{Version 4}    & 0.931  & 0.924  & 0.925  & 0.924  \\ 
\textbf{Version 5}    & 0.929  & 0.927  & 0.927  & 0.927  \\ 
\textbf{Version 6}    & 0.942  & 0.939  & 0.939  & 0.939  \\ 
\textbf{Version 7}    & 0.937  & 0.936  & 0.936  & 0.936  \\ 
\textbf{Version 8}    & 0.941  & 0.939  & 0.939  & 0.939  \\ 
\bottomrule
\end{tabular}
\end{table}

\section{Explainability of classification models}
\begin{figure}[ht]
  \centering
  \includegraphics[width=0.45\textwidth]{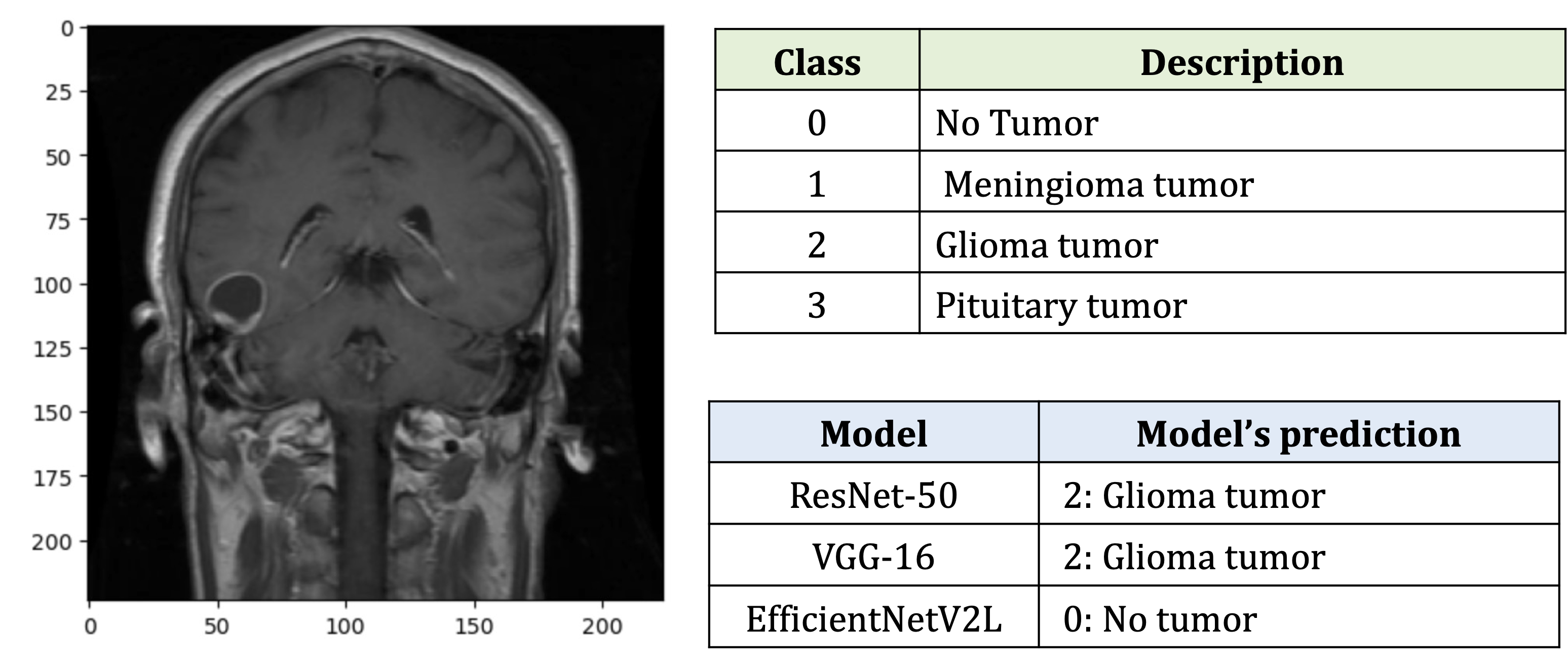}
  \captionsetup{justification=centering}
  \caption{Visualization samples and model predictions}
  \label{fig:data_info}
\end{figure}

In this section, we examine the difference in explainability based on classification performance and the difference between the three XAI techniques. To see the difference in explainability of models based on performance, we use ResNet-50 and version 2 of VGG-16 from Section 4.4 as good performers and EfficientNet from Section 4.1 as bad performers. The better performing models are visualized as correct and the worse performing models as incorrect samples. Here are the original photos in the sample and the model's predictions (Figure \ref{fig:data_info}).

\subsection{LIME}
\begin{figure}[ht]
  \centering
  \includegraphics[width=0.45\textwidth]{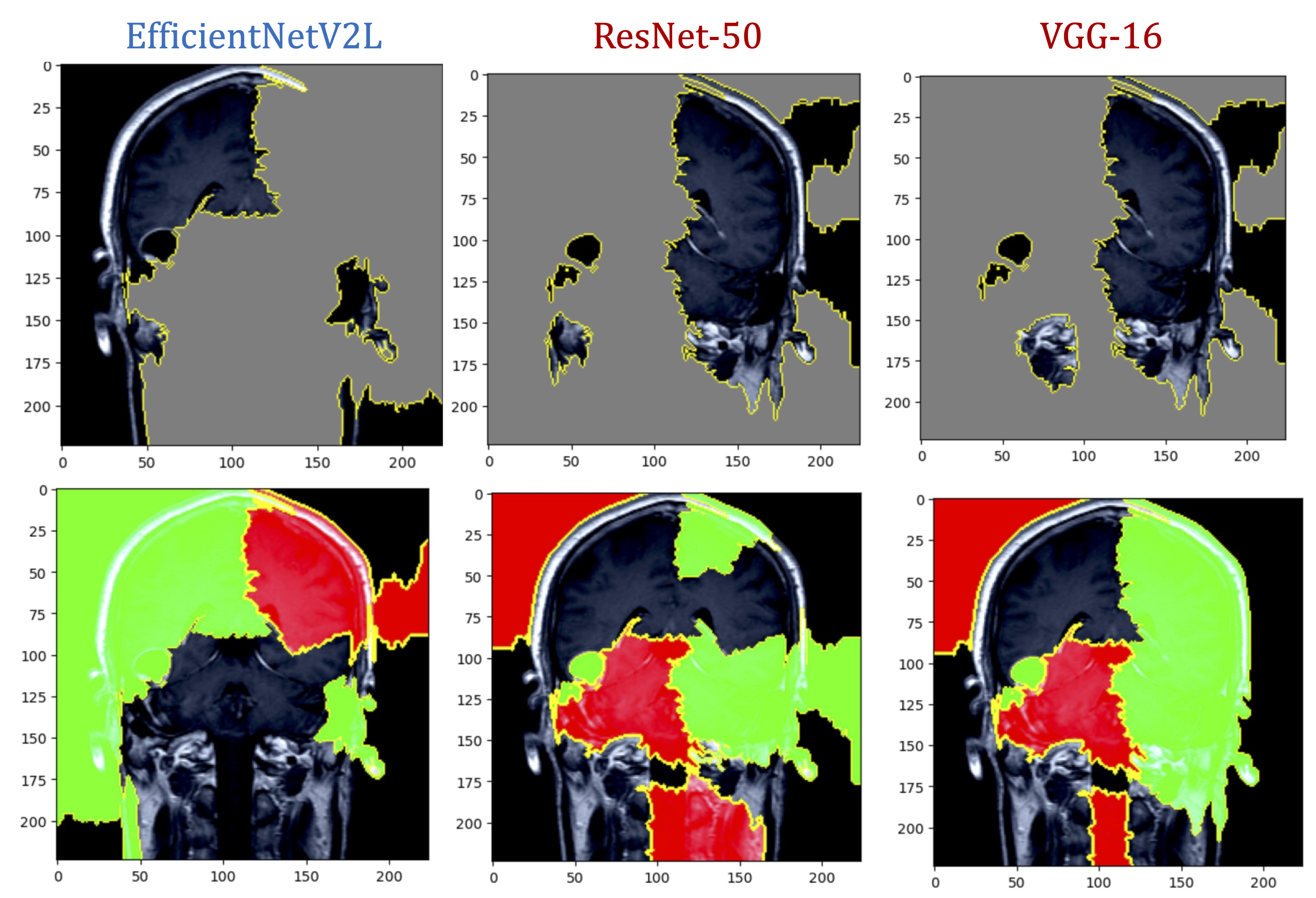}
  \captionsetup{justification=centering}
  \caption{LIME visualization result}
  \label{fig:LIME_exp}
\end{figure}

To visualize LIME, the LimeImageExplainer function from the LIME package was used. The feature selection is selected as auto, num\_samples=1000, num\_features=10, and visualized for Top Predicted Class. The top of the visualization shows only the super pixels, while the bottom shows the pixels that contribute negatively (red) and positively (green) (Figure \ref{fig:LIME_exp}).

\subsection{SHAP}
\begin{figure}[ht]
  \centering
  \includegraphics[width=0.45\textwidth]{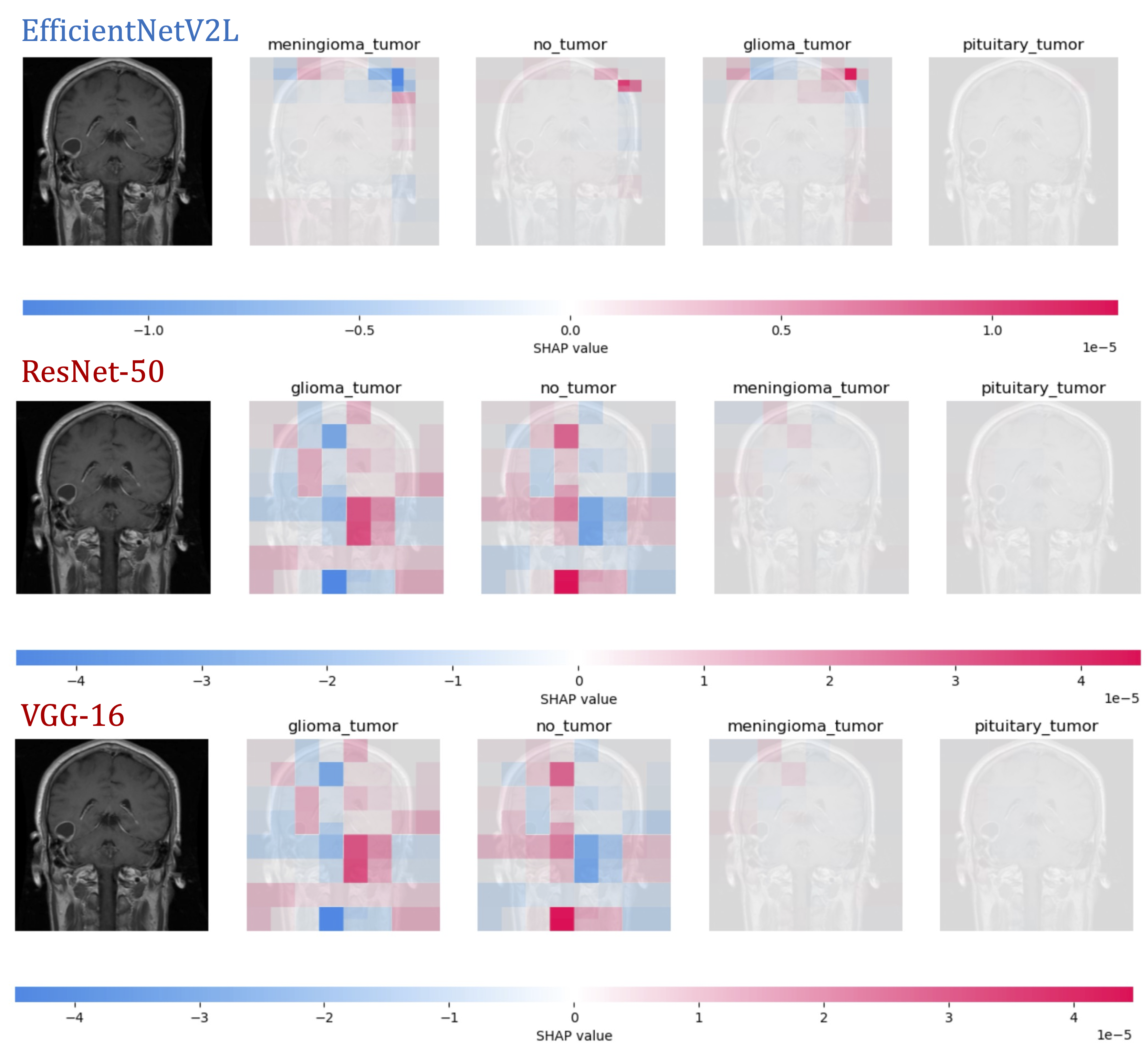}
  \captionsetup{justification=centering}
  \caption{SHAP visualization result}
  \label{fig:shap_exp}
\end{figure}

To visualize Shap Value,shap package was used. And, the classes were visualized sequentially, starting with those with the highest prediction probability. Positively contributing pixels are colored red, and negatively contributing pixels are colored blue (Figure \ref{fig:shap_exp}).

\subsection{Grad-CAM}
\begin{figure}[ht]
  \centering
  \includegraphics[width=0.45\textwidth]{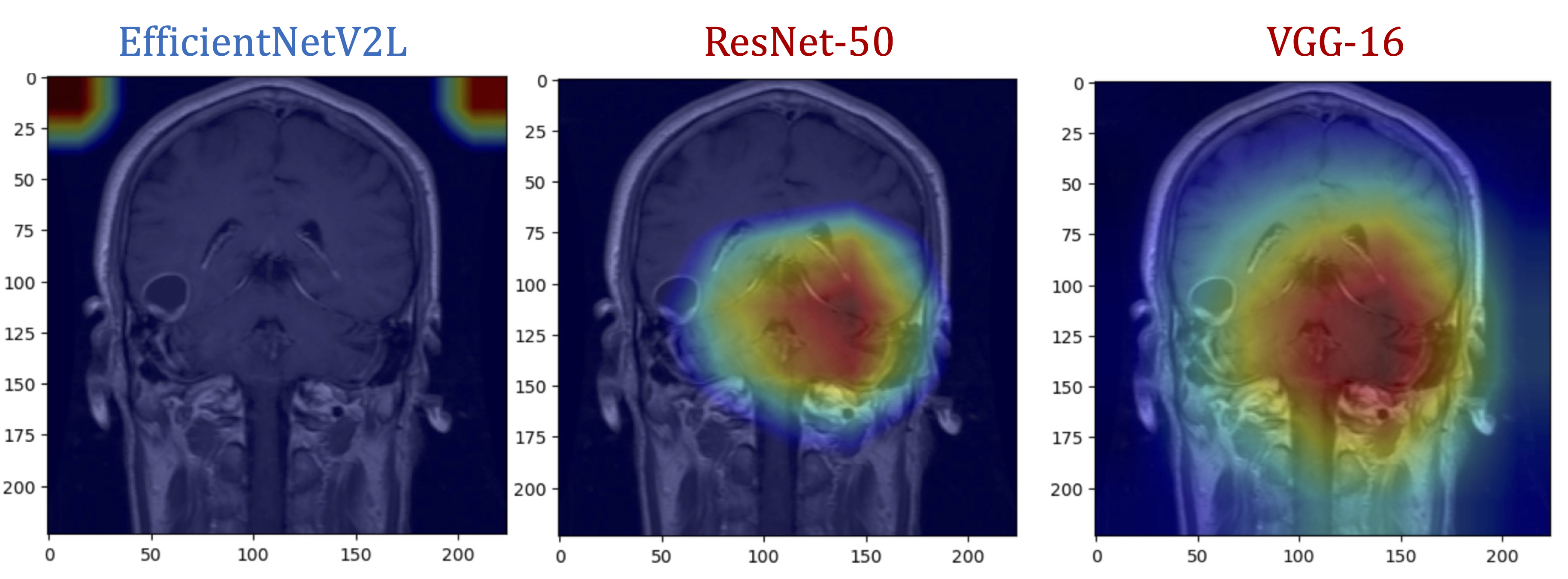}
  \captionsetup{justification=centering}
  \caption{Grad-CAM visualization result}
  \label{fig:CAM_result}
\end{figure}

To visualize the Grad-cam, we used GradientTape to compute the loss for the class, compute the output of the layer and the guided gradient, and generate a CAM based on that. We then overlay the Grad-CAM on top of the original image, convert it to a heatmap, and apply a weighted sum to the original image to create a visually enhanced image. To visualize the Grad-CAM, we implement a direct function. The visualized layer of the Grad-CAM is set to the last convolution layer, the same for all models. Here's the visualized result (Figure \ref{fig:CAM_result}).

\subsection{Impact of Accuracy}
For the good performing models, VGG-16 and ResNet-50, all three methods point to the right, which is where the brain tumour is located. However, for EfficientNet, we can see that LIME points to the left, SHAP points to the outer region of the brain, and Grad-CAM points to the area without the brain. Looking at these results, we can see that the regions visualised by the different XAI methods are slightly different, but they point to similar areas, indicating a good approximation of the reasoning behind the model's decision. We can also see that the decision boundaries are slightly different for the lower-performing models, presumably because the model made an incorrect guess, so the probability value for that class is low, and there are no regions that significantly helped the model make the decision.

\subsection{Comparison of XAI Methods}
In this section, we'll check the effectiveness of the visualization methods of the three methods. The three methods use different visualization methods. LIME visualizes the regions that influence the model's prediction decisions at the pixel level, while SHAP visualizes the degree of influence of each pixel in a bar chart. Grad-CAM displays a heatmap of which regions were paid attention to to make predictions, with darker colors indicating more importance. For LIME, it isolates regions that are SuperPixels, or shows positive and negative regions in green/red colors. SHAP uses heatmaps as a visualization method, while Grad-CAM shows a visualization similar to an Attention map.
For LIME, the super pixel visualization pulls out the important parts and shows them separately, so you don't have the problem of obscuring the original image like heat maps and attention maps. For the positive visualization of LIME and the visualization of SHAP and grad-cam, the original image is obscured. Therefore, the heatmap may cause clinicians to miss important visual details, and it may be necessary to provide the visualization with the original image for decision-making. In particular, for SHAP, the decision boundaries in the heatmap are in small blocks, and showing fine decision boundaries may not be suitable when it is necessary. In addition, the original image is blurred in the visualization, which may also increase visual fatigue.
For these reasons, when looking at these methods, the researchers believe that the best visualization method is LIME's super pixel visualization, and that LIME's positive and negative visualization and Grad-CAM's visualization method may be good when shown with the original image. Especially in the case of brain tumor diagnosis, the diagnosis can only be made visually, and since doctors have to review data from many patients, it is important that the explanation is provided in a way that does not increase visual fatigue and helps them make decisions. For this reason, the effectiveness of these methods was reviewed from a User eXperience (UX) perspective, but the researchers were UX experts and AI and vision experts without a medical background, so it may be difficult to verify the effectiveness of visualization methods suitable for the medical field.

\section{Conclusions}
We compare the performance of brain tumor detection by performing transfer learning using four representative models based on CNN and Transformer. We also check the effect of these methodologies on the performance of the model by applying two data augmentations and tuning hyper parameters such as learning rate, optimizer, and epochs. Furthermore, we check whether the performance of the model is improved by adding layers and dropouts in the pre-trained model. In this study, we found that data augmentation was not effective, but hyperparameter tuning and additional modeling had a direct impact on the performance of the model. This allows us to propose a model that can robustly detect brain tumors even with a small dataset and how to improve its performance.\\
In addition, to ensure the explainability of the developed models, we apply representative XAI methodologies LIME, SHAP, and Grad-CAM, and check the visualization effect between the methods and confirm the difference in explainability between high- and low-performing models. This is expected to increase the transparency of the models and provide insights into key features of brain tumor detection. This may encourage clinical adoption of these models by strengthening the confidence of healthcare professionals.\\
Although this study used different baseline models, performed image preprocessing and parameter tuning, additional modeling, and applied different XAI methodologies, there are still some limitations. First, two augmentation methods were examined and showed no performance benefit, which may indicate that the augmentation method was not appropriate for the dataset, or that the dataset was such that the effectiveness of the augmentation was negligible. Second, we used different baseline models and experimental conditions, but there are still models and parameters that can be experimented with. Third, in order to verify the effectiveness of XAI, human evaluation may be needed to verify the effectiveness of the three methodologies by recruiting real decision makers, medical staff. Therefore, future studies can improve the completeness of the study by applying various aumgentation methods, diversifying model and parameter conditions, and human evaluation.\\

\appendix
\section{Appendix}
\begin{figure}[ht]
  \centering
  \includegraphics[width=0.45\textwidth]{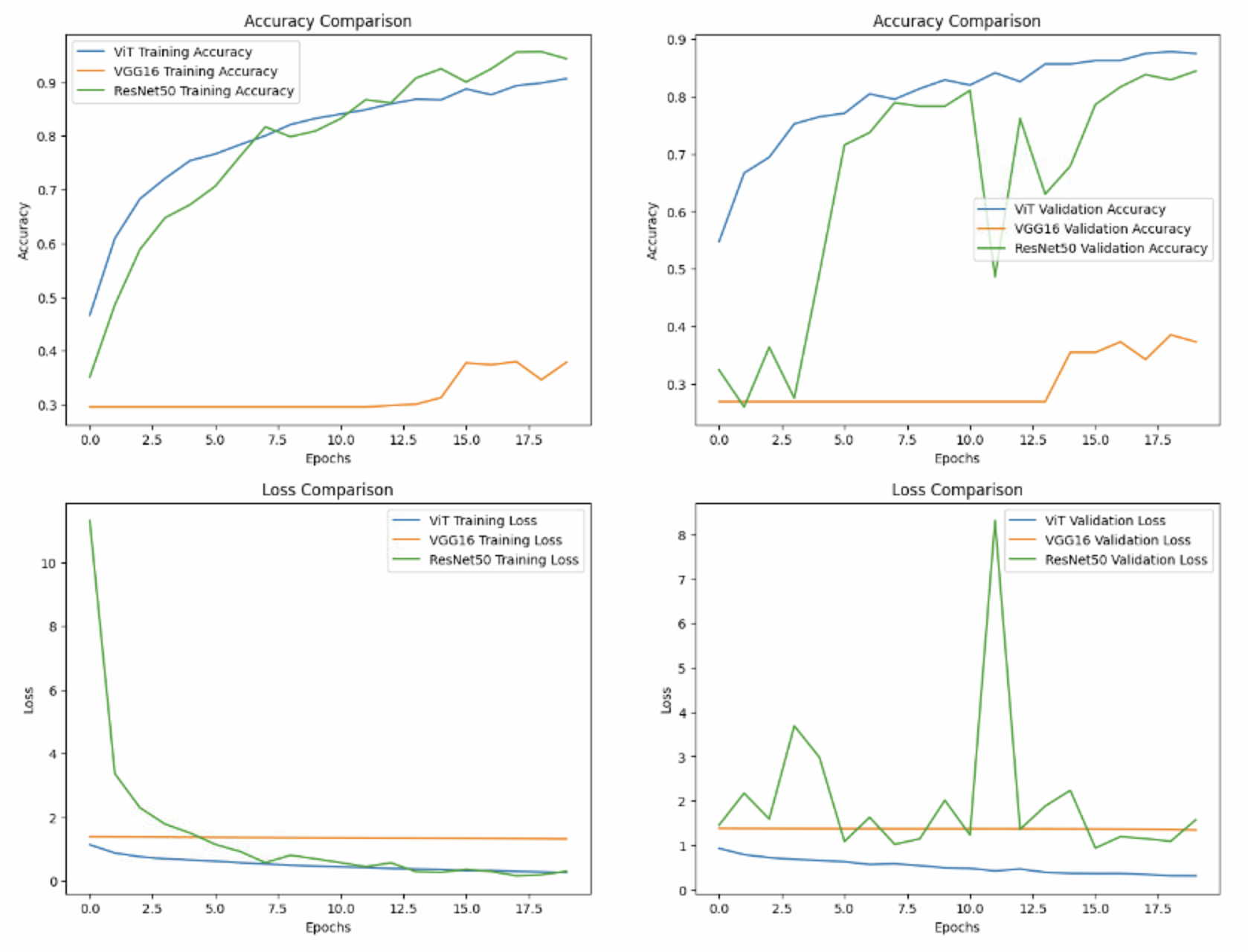}
  \caption{Performance of Not Pretrained Model}
  \label{fig:Not_pretrained}
\end{figure}

This result compares the performance of the model for weights that are not pretrained (Figure-\ref{fig:Not_pretrained}, Table-\ref{tab:not_pretrained performance}). We excluded EfficientNet and set the number of epochs to 20 because it requires a lot of computing power and was not feasible to experiment with in our local environment.
Using transfer learning, or not using it, may or may not be beneficial depending on the model. However, training a model from scratch requires significantly more computing resources than using transfer learning, which can increase the barrier to entry for AI applications. In the case of ViT, it was found that the performance decreased after transfer learning, but in the case of ResNet-50 and VGG-16, even if only a few epochs were trained during transfer learning, they showed high accuracy, and in the case of EfficientNetV2L, it was not possible to learn with zero computing resources. Therefore, it is recommended to prioritize transfer learning and improve the method of using non-pre-trained models if the performance is not excellent like ViT. 
\begin{table}[ht]
\centering
\caption{Performance of Not-pretrained Models}
\label{tab:not_pretrained performance}
\begin{tabular}{l c c c}
\toprule
          & \textbf{ResNet} & \textbf{VGG-16} & \textbf{ViT}   \\ 
\midrule
\textbf{Precision} & 0.847  & 0.345 & 0.912 \\ 
\textbf{Recall}    & 0.803  & 0.372 & 0.906 \\ 
\textbf{F1-score}  & 0.808  & 0.284 & 0.907 \\ 
\textbf{Accuracy}  & 0.819  & 0.390 & 0.908 \\ 
\bottomrule
\end{tabular}
\end{table}

{\small
\bibliographystyle{ieee_fullname}
\bibliography{egbib}

\begin{thebibliography}{10}\itemsep=-1pt

\bibitem{araujo2020dr}
Teresa Araujo, Guilherme Aresta, Lu{\'\i}s Mendon{\c{c}}a, Susana Penas, Carolina Maia, {\^A}ngela Carneiro, Ana~Maria Mendon{\c{c}}a, and Aur{\'e}lio Campilho.
\newblock Dr| graduate: Uncertainty-aware deep learning-based diabetic retinopathy grading in eye fundus images.
\newblock {\em Medical Image Analysis}, 63:101715, 2020.

\bibitem{dosovitskiy2020image}
Alexey Dosovitskiy, Lucas Beyer, Alexander Kolesnikov, Dirk Weissenborn, Xiaohua Zhai, Thomas Unterthiner, Mostafa Dehghani, Matthias Minderer, Georg Heigold, Sylvain Gelly, et~al.
\newblock An image is worth 16x16 words: Transformers for image recognition at scale.
\newblock {\em arXiv preprint arXiv:2010.11929}, 2020.

\bibitem{fidel2020explainability}
Gil Fidel, Ron Bitton, and Asaf Shabtai.
\newblock When explainability meets adversarial learning: Detecting adversarial examples using shap signatures.
\newblock In {\em 2020 international joint conference on neural networks (IJCNN)}, pages 1--8. IEEE, 2020.

\bibitem{grover2023braintumor}
Pratham Grover.
\newblock Brain tumor classification.
\newblock \url{https://www.kaggle.com/datasets/prathamgrover/brain-tumor-classification}, 2023.

\bibitem{he2016deep}
Kaiming He, Xiangyu Zhang, Shaoqing Ren, and Jian Sun.
\newblock Deep residual learning for image recognition.
\newblock In {\em Proceedings of the IEEE conference on computer vision and pattern recognition}, pages 770--778, 2016.

\bibitem{hossain2023vision}
Shahriar Hossain, Amitabha Chakrabarty, Thippa~Reddy Gadekallu, Mamoun Alazab, and Md~Jalil Piran.
\newblock Vision transformers, ensemble model, and transfer learning leveraging explainable ai for brain tumor detection and classification.
\newblock {\em IEEE Journal of Biomedical and Health Informatics}, 2023.

\bibitem{joshi2022forged}
Raunak Joshi, Abhishek Gupta, Nandan Kanvinde, and Pandharinath Ghonge.
\newblock Forged image detection using sota image classification deep learning methods for image forensics with error level analysis.
\newblock In {\em 2022 13th International Conference on Computing Communication and Networking Technologies (ICCCNT)}, pages 1--6. IEEE, 2022.

\bibitem{kim2020understanding}
Byung-Hoon Kim and Jong~Chul Ye.
\newblock Understanding graph isomorphism network for rs-fmri functional connectivity analysis.
\newblock {\em Frontiers in neuroscience}, 14:630, 2020.

\bibitem{kubach2020same}
Joshua Kubach, Angelika Muhlebner-Fahrngruber, Figen Soylemezoglu, Hajime Miyata, Pitt Niehusmann, Mrinalini Honavar, Fabio Rogerio, Se-Hoon Kim, Eleonora Aronica, Rita Garbelli, et~al.
\newblock Same same but different: A web-based deep learning application revealed classifying features for the histopathologic distinction of cortical malformations.
\newblock {\em Epilepsia}, 61(3):421--432, 2020.

\bibitem{lundberg2017unified}
Scott~M Lundberg and Su-In Lee.
\newblock A unified approach to interpreting model predictions.
\newblock {\em Advances in neural information processing systems}, 30, 2017.

\bibitem{panati2022feature}
Chandana Panati, Simon Wagner, and Stefan Br{\"u}ggenwirth.
\newblock Feature relevance evaluation using grad-cam, lime and shap for deep learning sar data classification.
\newblock In {\em 2022 23rd International Radar Symposium (IRS)}, pages 457--462. IEEE, 2022.

\bibitem{ribeiro2016should}
Marco~Tulio Ribeiro, Sameer Singh, and Carlos Guestrin.
\newblock " why should i trust you?" explaining the predictions of any classifier.
\newblock In {\em Proceedings of the 22nd ACM SIGKDD international conference on knowledge discovery and data mining}, pages 1135--1144, 2016.

\bibitem{selvaraju2017grad}
Ramprasaath~R Selvaraju, Michael Cogswell, Abhishek Das, Ramakrishna Vedantam, Devi Parikh, and Dhruv Batra.
\newblock Grad-cam: Visual explanations from deep networks via gradient-based localization.
\newblock In {\em Proceedings of the IEEE international conference on computer vision}, pages 618--626, 2017.

\bibitem{shatnawi2023deep}
Maad Shatnawi, Frdoos Albreiki, Ashwaq Alkhoori, and Mariam Alhebshi.
\newblock Deep learning and vision-based early drowning detection.
\newblock {\em Information}, 14(1):52, 2023.

\bibitem{simonyan2014very}
Karen Simonyan and Andrew Zisserman.
\newblock Very deep convolutional networks for large-scale image recognition.
\newblock {\em arXiv preprint arXiv:1409.1556}, 2014.

\bibitem{tammina2019transfer}
Srikanth Tammina.
\newblock Transfer learning using vgg-16 with deep convolutional neural network for classifying images.
\newblock {\em International Journal of Scientific and Research Publications (IJSRP)}, 9(10):143--150, 2019.

\bibitem{theckedath2020detecting}
Dhananjay Theckedath and RR Sedamkar.
\newblock Detecting affect states using vgg16, resnet50 and se-resnet50 networks.
\newblock {\em SN Computer Science}, 1:1--7, 2020.

\bibitem{tu2021gold}
Li Tu, Zheng Luo, Yun-Long Wu, Shuaidong Huo, and Xing-Jie Liang.
\newblock Gold-based nanomaterials for the treatment of brain cancer.
\newblock {\em Cancer Biology \& Medicine}, 18(2):372, 2021.

\bibitem{xu2021forest}
Renjie Xu, Haifeng Lin, Kangjie Lu, Lin Cao, and Yunfei Liu.
\newblock A forest fire detection system based on ensemble learning.
\newblock {\em Forests}, 12(2):217, 2021.

\bibitem{yadav2019deep}
Samir~S Yadav and Shivajirao~M Jadhav.
\newblock Deep convolutional neural network based medical image classification for disease diagnosis.
\newblock {\em Journal of Big data}, 6(1):1--18, 2019.

\bibitem{yan2021neural}
Jiun-Lin Yan, Cheng-Hong Toh, Li Ko, Kuo-Chen Wei, and Pin-Yuan Chen.
\newblock A neural network approach to identify glioblastoma progression phenotype from multimodal mri.
\newblock {\em Cancers}, 13(9):2006, 2021.

\bibitem{zhu2021interpreting}
Manli Zhu, Qianhui Men, Edmond~SL Ho, Howard Leung, and Hubert~PH Shum.
\newblock Interpreting deep learning based cerebral palsy prediction with channel attention.
\newblock In {\em 2021 IEEE EMBS International Conference on Biomedical and Health Informatics (BHI)}, pages 1--4. IEEE, 2021.

\bibitem{zhu2019guideline}
Peifei Zhu and Masahiro Ogino.
\newblock Guideline-based additive explanation for computer-aided diagnosis of lung nodules.
\newblock In {\em Interpretability of Machine Intelligence in Medical Image Computing and Multimodal Learning for Clinical Decision Support: Second International Workshop, iMIMIC 2019, and 9th International Workshop, ML-CDS 2019, Held in Conjunction with MICCAI 2019, Shenzhen, China, October 17, 2019, Proceedings 9}, pages 39--47. Springer, 2019.

\end{thebibliography}
}

\end{document}